%% file: main.tex
\documentclass[]{bytedance}
\usepackage[toc,page,header]{appendix}

\usepackage{minitoc}
\usepackage{amsfonts}
\usepackage{amssymb}
\usepackage{tabularx}
\usepackage{listings}
\usepackage{xcolor}
\usepackage{cancel}

\usepackage{tabulary,multirow,xspace}
\usepackage{fixmath,mathtools,nicefrac,mmstyle}
\usepackage{subcaption}
\usepackage{caption}
\usepackage{wrapfig} 
\usepackage[misc]{ifsym} 
\usepackage{colortbl}

\usepackage{wrapfig}
\usepackage{multicol}
\usepackage[most]{tcolorbox}
\usepackage{pifont}

\definecolor{codegreen}{rgb}{0,0.6,0}
\definecolor{codegray}{rgb}{0.5,0.5,0.5}
\definecolor{codepurple}{rgb}{0.58,0,0.82}
\definecolor{backcolour}{rgb}{0.95,0.95,0.92}
\definecolor{boxblue}{RGB}{57,89,163}
\definecolor{boxbluebg}{RGB}{230,237,250} 

\lstdefinestyle{mystyle}{
    backgroundcolor=\color{backcolour},   
    commentstyle=\color{codegreen},
    keywordstyle=\color{magenta},
    numberstyle=\tiny\color{codegray},
    stringstyle=\color{codepurple},
    basicstyle=\ttfamily\footnotesize,
    breakatwhitespace=false,         
    breaklines=true,                 
    captionpos=b,                    
    keepspaces=true,                 
    numbers=none,                    
    numbersep=5pt,                  
    showspaces=false,                
    showstringspaces=false,
    showtabs=false,                  
    tabsize=2
}
\definecolor{mygray1}{gray}{.95}
\definecolor{mygray2}{gray}{.9}
\definecolor{mygray3}{gray}{.95}
\usepackage{pifont}

\newlength\savewidth
\newcolumntype{x}[1]{>{\centering\arraybackslash}p{#1pt}}

\newcommand{\app}{\raise.17ex\hbox{$\scriptstyle\sim$}}

\usepackage{xcolor}
\input{macros}

\usepackage{listings}

\definecolor{carnelian}{rgb}{0.7, 0.11, 0.11}
\usepackage{booktabs}
\usepackage{makecell}
\usepackage{multirow}

\usepackage{adjustbox}
\usepackage{cuted} 

\usepackage{algorithm}
\usepackage{algorithmic}
\definecolor{myblue}{RGB}{210, 225, 255}
\definecolor{mytextblue}{RGB}{51, 161, 201}
\definecolor{mypurple}{RGB}{218, 112, 214}

\definecolor{commentgreen}{rgb}{0.1, 0.4, 0.1}
\definecolor{keywordblue}{rgb}{0.1, 0.1, 0.7}
\definecolor{stringred}{rgb}{0.7, 0.1, 0.1}

\lstdefinestyle{mystyle}{
    commentstyle=\color{commentgreen},
    keywordstyle=\color{keywordblue},   
    stringstyle=\color{stringred},
    basicstyle=\ttfamily\scriptsize, 
    breaklines=true,
    keepspaces=true,
    showstringspaces=false,
    frame=none,                     
    language=Python, 
}

\title{DreamStyle: A Unified Framework for Video Stylization}

\author{
\centerline{
Mengtian Li $^{\dagger}$ \quad
Jinshu Chen \quad
Songtao Zhao $^{\ddagger}$ \quad
Wanquan Feng
}
\centerline{
Pengqi Tu \quad
Qian He
}
}

\affiliation[]{Intelligent Creation Lab, ByteDance}

\input{sections/0_abstract}
\date{\today}
\checkdata[Project Page]{\url{https://lemonsky1995.github.io/dreamstyle/}}

\begin{document}
\maketitle

\renewcommand{\thefootnote}{\fnsymbol{footnote}}
\footnotetext{$\dagger$ Corresponding author, $\ddagger$ Project lead.}
\renewcommand{\thefootnote}{\arabic{footnote}}

\input{sections/1_introduction}
\input{sections/2_related_work}
\input{sections/3_method}
\input{sections/4_experiments}
\input{sections/5_conclusion}

\bibliographystyle{ieeenat_fullname}
\bibliography{main}

\end{document}

%% file: macros.tex
\usepackage{graphicx}
\usepackage{amssymb}
\usepackage{pifont}
\usepackage{floatrow}
\usepackage{amsmath} 
\usepackage{float}
\usepackage{wrapfig}
\usepackage{multirow}
\usepackage{tcolorbox}
\tcbuselibrary{breakable, skins, raster}
\usepackage{listings}

%% file: sections/0_abstract.tex
\abstract{
Video stylization, an important downstream task of video generation models, has not yet been thoroughly explored. Its input style conditions typically include text, style image, and stylized first frame. Each condition has a characteristic advantage: text is more flexible, style image provides a more accurate visual anchor, and stylized first frame makes long-video stylization feasible. However, existing methods are largely confined to a single type of style condition, which limits their scope of application. Additionally, their lack of high-quality datasets leads to style inconsistency and temporal flicker. To address these limitations, we introduce \textbf{DreamStyle}, a unified framework for video stylization, supporting (1) text-guided, (2) style-image-guided, and (3) first-frame-guided video stylization, accompanied by a well-designed data curation pipeline to acquire high-quality paired video data. DreamStyle is built on a vanilla Image-to-Video (I2V) model and trained using a Low-Rank Adaptation (LoRA) with token-specific up matrices that reduces the confusion among different condition tokens. Both qualitative and quantitative evaluations demonstrate that DreamStyle is competent in all three video stylization tasks, and outperforms the competitors in style consistency and video quality.
}

%% file: sections/1_introduction.tex
\section{Introduction}
\label{sec:intro}

\begin{figure}[t]
    \centering
    \includegraphics[width=1.0\linewidth]{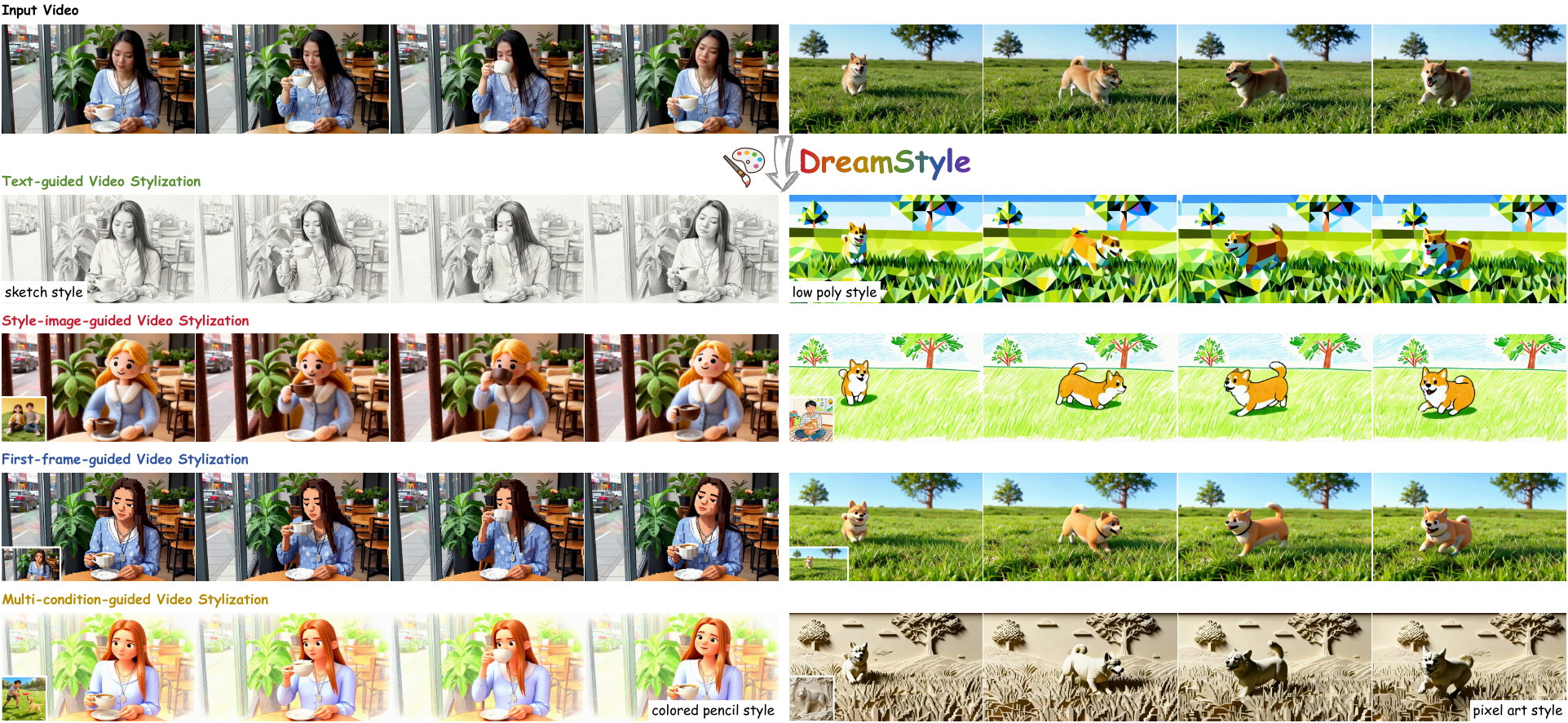}
    \caption{We propose DreamStyle, a unified video stylization framework, which provides a flexible and practical tool for users to create high-quality stylized videos. Given an input video and the reference styles in forms of text, style image, or stylized first frame, DreamStyle faithfully generates videos that align with the desired style—while preserving the main content of the input video.}
    \label{fig:teaser}
\end{figure}

Video stylization stands as a compelling yet challenging task in the field of visual content generation.
Existing video stylization approaches are confronted with the following critical limitations:
\textbf{(1) Limited stylization capabilities due to single-modality condition.} Text prompts and style images are the two dominant style conditions, but both suffer from inherent flaws. Text prompts are typically ambiguous and unconstrained, failing to precisely describe most abstract styles. Style images, while more visually accurate, exhibit inferior user-friendliness, flexibility, and creativity—it is difficult to acquire a suitable style image, especially for unseen styles. Consequently, most existing methods are confined to styles that are either explicitly describable via text or have clear visual references, exhibiting limited generalization to novel styles.
\textbf{(2) Scarcity of high-quality modality-aligned training data.} Some existing methods~\cite{ye2025stylemaster,chefer2024still,liu2024stylecrafter} acquire stylization capabilities from image stylization datasets and subsequently generalize to the video domain assisted by a pre-trained video generation model. This paradigm inherently introduces an unavoidable trade-off among style consistency, temporal consistency, and motion dynamics. More recently, UNIC~\cite{ye2025unic} synthesizes stylized videos via a Text-to-Video (T2V) model and employs a gray tiled ControlNet to invert these stylized videos into their realistic counterparts, thereby constructing paired video data. However, its stylization quality is limited by the T2V model and it fails to handle the styles involving geometric deformation due to the strict alignment of tile ControlNet.
\textbf{(3) Insufficient exploration of potential extended applications.} Current research predominantly focuses on basic stylization capabilities, with limited attention to high-demand extended scenarios, such as multi-style fusion and long-video stylization.

To tackle the aforementioned challenges, we propose \textbf{DreamStyle}, which includes the following three key innovations:
First, we introduce a unified Video-to-Video (V2V) stylization framework, which is built upon a vanilla I2V model. Through a meticulous design of the condition injection mechanism, we manage to unify diverse forms of style guidance including text prompt, style image, and stylized first frame into a single model, extending the I2V base model to V2V domain while preserving its original architecture and inherent capabilities. We further employ a modified LoRA module composed of a shared down matrix and token-specific up matrices to enhance the multi-task adaptability.
Second, we present a systematic data curation pipeline tailored for the video stylization task, the core of which involves two steps: (1) stylizing the initial frame of a real-world video using image stylization techniques and (2) generating the full stylized video sequence from the stylized first frame via an I2V model equipped with ControlNets. To guarantee the data quality, we further adopt a hybrid filtering strategy consisting of automatic and manual filtering. Leveraging this pipeline, we construct two datasets with distinct scales and quality to facilitate multi-stage training.
Finally, comprehensive evaluations across multiple dimensions exhibit that our \textbf{unified} model, DreamStyle, achieves competitive performance against \textbf{specialized} models across various video stylization tasks. Notably, we also demonstrate that allowing multiple style conditions within a single forward process is a crucial design for improving the effectiveness and controllability of video stylization—such a design unlocks the model's capability to support more potential extended applications, such as multi-style fusion and long-video stylization.

Overall, our contributions are summarized as follows.

\textbf{Paradigm.} We introduce DreamStyle, which consists of a unified framework that supports text-guided, style-image-guided, and first-frame-guided video stylization; a well-designed pipeline for constructing high-quality paired data for video stylization.

\textbf{Technology.} DreamStyle framework presents a condition injection mechanism that enables seamless handling of diverse video stylization tasks within a unified model; a novel LoRA module that mitigates the interference among different condition tokens.

\textbf{Scalability.} DreamStyle data curation pipeline is practical and scalable for video stylization, overcoming the scarcity of high-quality data and the inherent trade-off between style fidelity and temporal coherence.

\textbf{Significance.} DreamStyle outperforms specialized competitors in various video stylization tasks and exhibits the potential for under-explored extended tasks.

%% file: sections/2_related_work.tex
\section{Related Work}
\label{sec:relwork}

\subsection{Video Diffusion Model}
Diffusion models~\cite{sohl2015deep, ho2020denoising, nichol2021improved, dhariwal2021diffusion, saharia2022photorealistic} have driven remarkable advancements in visual content generation. Latent Diffusion Models~\cite{rombach2022high, podell2024sdxl} (LDMs) further optimize this paradigm by training a diffusion network in the latent space of pretrained Variational Autoencoder~\cite{Kingma2014} (VAE) to reduce computational complexity, becoming the mainstream solution. Early video diffusion models~\cite{blattmann2023stable, blattmann2023align, guo2024animatediff, guo2024i2v} were mostly built upon pretrained image diffusion models (typically U-Net~\cite{ronneberger2015u} architectures), incorporating temporal modules to handle temporal consistency. However, their isolated processing of spatial and temporal information inherently limited their quality and consistency. With the release of Sora~\cite{brooks2024video} and its epoch-making generation quality, researchers notice the potential of Diffusion Transformer~\cite{peebles2023scalable} (DiT) for video generation. Recent DiT-based methods~\cite{hong2023cogvideo, yang2025cogvideox, wan2025wan, gao2025seedance, kong2024hunyuanvideo} apply a unified manner to model the video in spatial-temporal domain, and scale the capability of DiT by more parameters, data and computing resources, achieving more high-quality and consistent video generation.

\subsection{Image Stylization}
Gatys et al.~\cite{gatys2016image} pioneered image stylization using neural networks. Early methods~\cite{gatys2016image, chen2017stylebank, huang2017arbitrary, risser2017stable} relied on the statistical descriptors (such as gram matrices, mean and standard deviation, and histograms) extracted from a pretrained VGG~\cite{simonyan2015very} network to represent and transfer style information. However, due to the limitations of the capability of generation and style extraction, they could only achieve simple texture and color transfer, often resulting in suboptimal visual quality. Recent methods~\cite{ye2023ip, wang2024instantstyle, xing2024csgo, qi2024deadiff} benefit from the advances of diffusion models to improve basic quality, and CLIP~\cite{radford2021learning} to extract high-level semantic information from style images. StyleTokenizer~\cite{li2024styletokenizer} further improves the style extractor with contrastive learning using a self-collected style dataset. Given the importance of high-quality datasets for stylization, OmniStyle~\cite{wang2025omnistyle} constructs a large-scale paired dataset using six state-of-the-art (SOTA) image stylization methods, and leverages a unified DiT backbone to extract style features and generate images, yielding new SOTA performance.

\subsection{Video Stylization}
Extending image-based tasks to video domain is a major trend in current research, and stylization is no exception. TokenFlow~\cite{geyer2024tokenflow} and AnyV2V~\cite{ku2024anyvv} achieve video stylization by leveraging image stylization techniques to stylize the first frame or key frames and then propagate it to the entire video sequence. However, these approaches can not perform video stylization independently, and rely on a time-consuming DDIM~\cite{song2021denoising} inversion. UniVST~\cite{song2024univst} further DDIM inverses the style image and leverage AdaIN~\cite{huang2017arbitrary} to guide the denoising progress of noisy video by the inverted features of style. StyleCrafter~\cite{liu2024stylecrafter} utilizes CLIP to extract style features and inject these features into the denoising U-Net via dual cross-attention. More recently, StyleMaster~\cite{ye2025stylemaster} upgrades to DiT backbone and incorporates both global and local style extractors, resorting to StillMoving~\cite{chefer2024still} to train a LoRA~\cite{hu2022lora} for temporal attention to bridge the gap between image and video. However, this scheme requires explicit temporal modeling within the base model, which deviates from mainstream architectures. Moreover, a limitation shared by all these methods is their lack of stylized video datasets, resulting in suboptimal visual quality and temporal consistency.

%% file: sections/3_method.tex
\newpage
\section{Method}
\label{sec:method}

\subsection{Data Curation Pipeline}
\label{sec:method_data}
\begin{figure*}[t]
    \centering
    \includegraphics[width=1.0\linewidth]{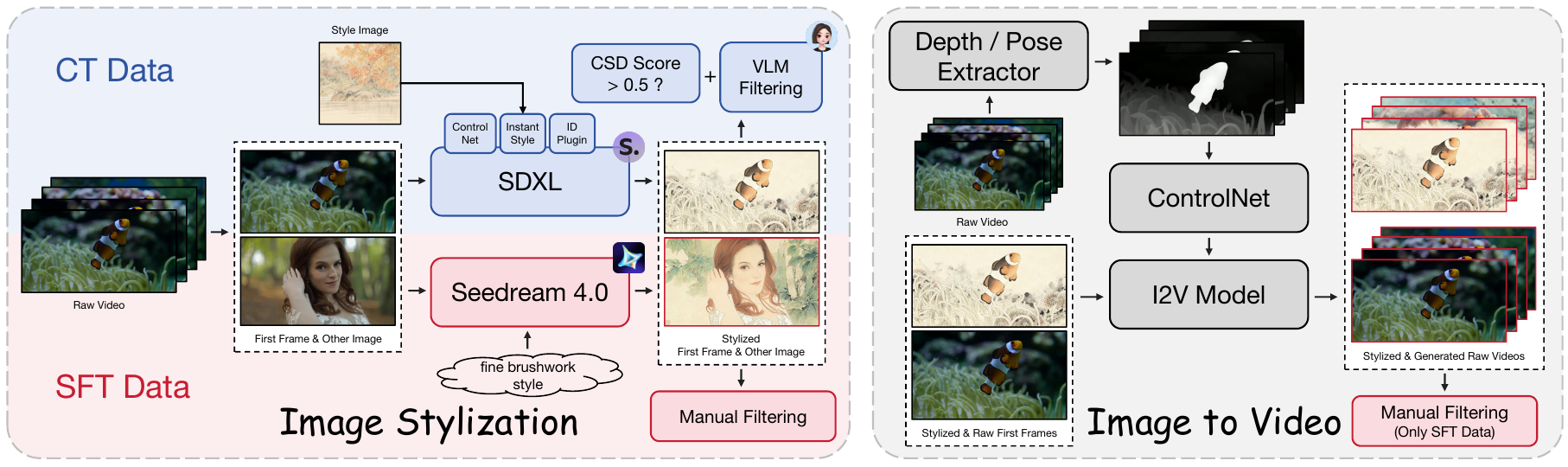}
    \caption{Data Curation Pipeline. We propose generating the training data with two key steps: image stylization followed by image to video. Considering the characteristics of different image stylization techniques, we construct a CT dataset and a SFT dataset, where SDXL (equipped with ControlNet, InstantStyle, and ID plugin) and Seedream 4.0 are selected as their stylization models, respectively. For image to video, we utilize ControlNets to enhance the motion consistency between the generated stylized and raw videos. To ensure the data quality, we additionally apply automatic filtering for CT data and manual filtering for SFT data.}
    \label{fig:data_pipeline}
\end{figure*}

Given the fact that current image generation / editing models are superior to the video counterpart in terms of visual quality, structure, aesthetics and text following, we propose generating stylized video datasets with two key steps: (1) leverage the SOTA image stylization models to stylize the first frame of raw video; (2) utilize the I2V model to generate stylized video from the stylized first frame. Our data curation pipeline is illustrated in Fig.~\ref{fig:data_pipeline}. It is noteworthy that a high-quality first frame serves as crucial cues (e.g., style constraints and content anchors) to improve the overall quality of the entire video generated by I2V model.

To obtain the high-fidelity stylized first frame, we select InstantStyle~\cite{wang2024instantstyle} and Seedream 4.0~\cite{seedream2025seedream} as our image stylization models, which are proficient in style-image-guided and text-guided stylization, respectively. InstantStyle is a SDXL~\cite{podell2024sdxl} plugin, which we further equip with a depth ControlNet~\cite{zhang2023adding} and ID plugin~\cite{guo2024pulid} to constrain the consistency of structure and face identity. It is worthy noting that the text-guided stylization model typically produces better visual quality and style consistency, while the style-image-guided stylization model allows us to generate images with greater style diversity. Thus, we construct two datasets: (1) a large-scale stylized dataset for Continual Training (CT) generated via InstantStyle to ensure the core video stylization capability and generalization of DreamStyle; (2) a small-scale higher-quality stylized dataset for Supervised Fine-Tuning (SFT) generated with Seedream 4.0 to elevate the upper bound of DreamStyle.

\begin{figure}[htb]
    \centering
    \includegraphics[width=0.7\linewidth]{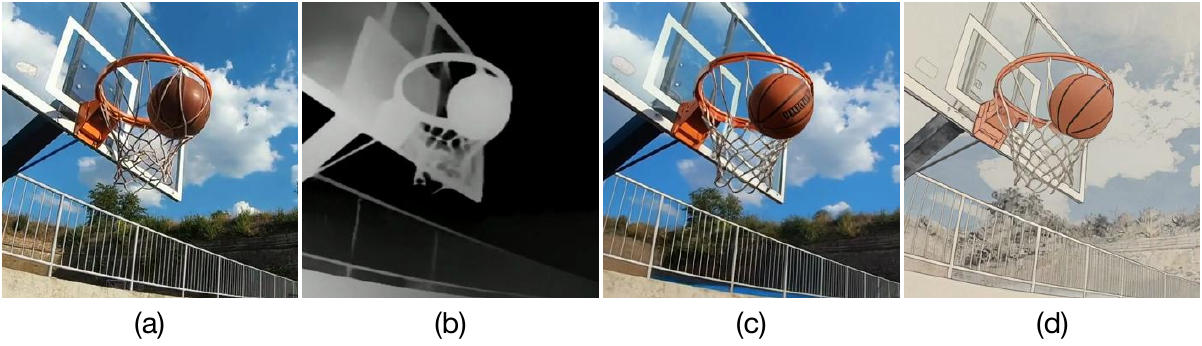}
    \caption{Example that depth fails to capture accurate detail. (a) The raw video frame, (b) the extracted depth map, (c) the generated realistic frame, (d) the generated stylized frame.}
    \label{fig:depth_example}
\end{figure}

It is critical to ensure the motion consistency between stylized video and raw video, so that we are able to construct stylized-raw video pairs. To this end, we customize two ControlNets (with control conditions of depth and human pose, respectively) for our in-house I2V model. The depth ControlNet is well-suited for general cases, while the human pose ControlNet offers a more precise control of human motion and especially allows for a larger deformation of driven objects without losing motion coherence. As illustrated in Fig.~\ref{fig:depth_example}, we observe that directly animating the stylized first frame using the control conditions extracted from the raw video and then making paired data proves to be suboptimal. This is because neither depth nor pose can fully capture the complex motion dynamics of the raw video, ultimately resulting in motion mismatches between stylized and raw videos. Thus, we adjust to utilizing the same control condition to drive the generation of both stylized and raw video frames, aiming to mitigate such mismatches.

\begin{figure*}[t]
    \centering
    \includegraphics[width=1.0\linewidth]{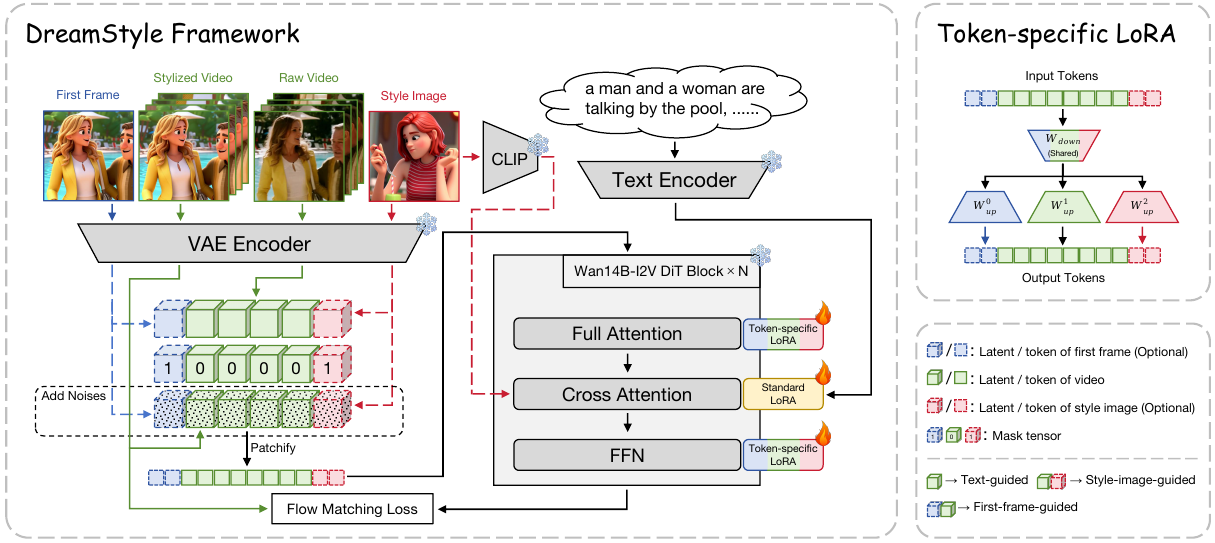}
    \caption{Overview of DreamStyle Framework. DreamStyle is built on the Wan14B-I2V model, integrating the text and raw-video conditions through the cross-attention and image channels of the base model, while the first-frame and style-image conditions serve as additional frames concatenated to the start and end of the frame sequence. We train it using a standard flow matching loss and a token-specific LoRA that contributes to distinguishing different condition tokens.}
    \label{fig:framework}
\end{figure*}

We formally denote our dataset as:
\begin{equation}
    \mathcal{D} = \{ (\mathbf{x}_i^{raw}, \mathbf{x}_i^{sty}, \mathbf{t}_i^{ns}, \mathbf{t}_i^{sty}, \mathbf{s}_i^{1...K}) | i=1,2,..,N \}
    \label{eq:dataset}
\end{equation}
where $\mathbf{x}_i^{raw}$ and $\mathbf{x}_i^{sty}$ are the raw and stylized videos, $\mathbf{t}_i^{ns}$ and $\mathbf{t}_i^{sty}$ are the text prompts that exclude / include style descriptions, and $\mathbf{s}_i^{1...K}$ denotes $K$ style reference images. To obtain $\mathbf{t}_i^{ns}$ and $\mathbf{t}_i^{sty}$, we utilize a Visual-Language Model~\cite{zhang2024vision} (VLM) to parse the stylized video $\mathbf{x}_i^{sty}$ and then generate the corresponding video caption. We restrict the VLM to exclude any style-related attributes (e.g., artistic genre, color palette, texture pattern, and hue) when generating $\mathbf{t}_i^{ns}$, so that $\mathbf{t}_i^{ns}$ contains only style-irrelevant descriptions. Regarding $\mathbf{s}_i^{1...K}$, we stylize $K$ additional images using the same guided condition (text prompt or style reference) as $\mathbf{x}_i^{sty}$. For the CT dataset, we further filter out those $\mathbf{s}_i^{1...K}$ with low style consistency detected by VLM and CSD~\cite{somepalli2024measuring} score, while we opt for manual filtering for the SFT dataset. Additionally, we manually verify the content consistency between each raw video and its stylized video in the SFT dataset. Finally, such two datasets enable our DreamStyle to support all three video stylization tasks.

\subsection{DreamStyle Framework}
As shown in Fig.~\ref{fig:framework}, our DreamStyle framework is built upon the Wan14B-I2V~\cite{wan2025wan} base model that incorporates additional image condition channels before the patchify layer. This design allows for the injection of raw video condition via these channels, rather than the in-context frames injection adopted in UNIC~\cite{ye2025unic}. A major advantage is that it involves minimal extra computational overhead, ensuring DreamStyle retains the efficiency of the original I2V model. Overall, we have four types of conditions to inject into the I2V model: the raw video along with three guided conditions of style, which are described in detail as follows.

\textbf{(1) Text condition.} We reuse the original textual cross-attention layers of Wan14B-I2V without introducing modifications. \textbf{(2) First-frame condition.} We feed the stylized first frame into the original image condition channels and set the mask channels of the first frame to $1.0$ in the same manner as the base model. \textbf{(3) Style-image condition.} Assuming that $\mathbf{z}^s\in \mathbb{R}^{C\times 1\times H\times W}$ (omit the subscript $i$) is the VAE encoded latent of the style reference image $\mathbf{s}_i^j$, we construct the final I2V model's input tensor for the style image via channel-wise concatenation:
\begin{equation}
    \mathbf{z}_t^s = add\_noise(\mathbf{z}^s, t) \oplus_c \mathbf{1}_{4\times 1\times H\times W} \oplus_c \mathbf{z}^s
\end{equation}
where $\oplus_c$ denotes the concatenation operation along the channel dimension, $add\_noise(\cdot,t)$ is the noise injection function of flow matching~\cite{lipmanflow} at timestep $t\in[0,1]$. The number of mask channels is $4$ in Wan14B-I2V, thus $\mathbf{1}_{4\times 1\times H\times W}$ represents a mask tensor filled with a constant value $1.0$. Wan14B-I2V also includes a native CLIP image feature branch, which we employ to inject high-level semantic features of the style reference image, thereby enhancing the consistency of style-related semantic information. \textbf{(4) Raw-video condition.} We first encode the raw video $\mathbf{x}_i^{raw}$ and stylized video $\mathbf{x}_i^{sty}$ to obtain their latents $\mathbf{z}^{raw}, \mathbf{z}^{sty}\in \mathbb{R}^{C\times F\times H\times W}$, and then channel-wise concatenate them with mask tensor $\mathbf{0}_{4\times F\times H\times W}$:
\begin{equation}
    \mathbf{z}_t^v = add\_noise(\mathbf{z}^{sty}, t) \oplus_c \mathbf{0}_{4\times F\times H\times W} \oplus_c \mathbf{z}^{raw}.
\end{equation}
Here we adopt the mask value $0.0$, following the principle of minimal modification of the base model. For the style-image-guided mode, $\mathbf{z}_t^s$ is treated as an additional frame and concatenated to the end of $\mathbf{z}_t^v$ via frame-wise concatenation $\oplus_f$: $\mathbf{z}_t^v \oplus_f \mathbf{z}_t^s$, enabling the model to incorporate the style-image condition. Similarly, for the first-frame-guided mode, the first-frame tensor $\mathbf{z}_t^{1st}$ is concatenated to the beginning of $\mathbf{z}_t^v$ via frame-wise concatenation: $\mathbf{z}_t^{1st} \oplus_f \mathbf{z}_t^v$.

To retain the base model's inherent generative capabilities, we adopt LoRA to train our DreamStyle. After patchification, the three conditions, first-frame, style-image, and raw-video, are transformed into their corresponding token sequences. However, these tokens serve distinct semantic roles, thus using a standard LoRA leads to inter-token confusion. Inspired by HydraLoRA~\cite{tian2024hydralora}, we propose adopting a modified LoRA with token-specific up matrices in full attention and feedforward (FFN) layers. That means, for an input token $\mathbf{x}_{in}$, we first project it using a shared down matrix $\mathbf{W}_{down}$, and then compute the output residual token $\mathbf{x}_{out} = \mathbf{W}_{up}^i \mathbf{W}_{down} \mathbf{x}_{in}$ with a specific up matrix $\mathbf{W}_{up}^i$ according to the token type $i\in\{0,1,2\}$, which is analogous to a LoRA MoE~\cite{dou2024loramoe} with manual routing. Such a LoRA enables the model to learn adaptive features tailored to the three types of tokens, and still be trained stably due to the large proportion of shared parameters.

\subsection{Training}
We follow the same optimization objective as flow matching to train our DreamStyle. Formally, we denote our model as $\mathbf{v}_\theta$, a function with five inputs: the video tensor $\mathbf{z}_t^v$, the timestep $t$, the first-frame tensor $\mathbf{z}_t^{1st}$, the style image tensor $\mathbf{z}_t^s$ and the text prompt $\mathbf{t}^{ns/sty}$. In each training batch, we randomly sample style conditions according to predefined ratios, thus the training objective is:
\begin{equation}
    \mathcal{L}(\theta) = \left\{
    \begin{aligned}
        & \mathbb{E}_\mathcal{D}\Vert \mathbf{v}_\theta(\mathbf{z}_t^v,t,\emptyset,\emptyset,\mathbf{t}^{sty}) -(\mathbf{z}^{sty}-\epsilon) \Vert^2 & (\mathrm{I}) \\
        & \mathbb{E}_\mathcal{D}\Vert \mathbf{v}_\theta(\mathbf{z}_t^v,t,\emptyset,\mathbf{z}_t^s,\mathbf{t}^{ns}) -(\mathbf{z}^{sty}-\epsilon) \Vert^2 & (\mathrm{II}) \\
        & \mathbb{E}_\mathcal{D}\Vert \mathbf{v}_\theta(\mathbf{z}_t^v,t,\mathbf{z}_t^{1st},\emptyset,\mathbf{t}^{ns}) -(\mathbf{z}^{sty}-\epsilon) \Vert^2 & (\mathrm{III})
    \end{aligned}
    \right.
\end{equation}
where $\epsilon\in\mathcal{N}(0,1)$ is a Gaussian noise and $(\mathrm{I})\sim(\mathrm{III})$ correspond to the loss terms for the text-guided, the style-image-guided and the first-frame-guided tasks, respectively. As mentioned in Sec.~\ref{sec:method_data}, we make two datasets with different scales and quality, thus adopting a two-stage training strategy. In the first stage, we train DreamStyle on the large-scale CT dataset, allowing the model to learn diverse styles and establish a foundational capability to handle all three style conditions. In the second stage, a higher-quality SFT dataset is used to further finetune DreamStyle, aiming to improve visual quality and style consistency.

%% file: sections/4_experiments.tex
\newpage
\section{Experiments}
\label{sec:exp}

\subsection{Implementation Details}
Through the data curation pipeline (Sec.~\ref{sec:method_data}), we construct $40$K and $5$K stylized-raw video pairs for the CT stage and SFT stage training, where the video resolution is 480P and the length is up to $81$ frames. In the CT dataset, each sample includes exactly one style reference while the samples from the SFT dataset contain $1 \sim 16$ style images, with one randomly selected for training. During the training, we empirically set the sampling ratio of the three style conditions (text-guided, style-image-guided and first-frame-guided) to $1:2:1$. The training process is performed on NVIDIA GPUs, with each GPU accommodating a per-GPU batch-size of $1$. To stabilize training, we further adopt a 2-step gradient accumulation strategy, resulting in a larger effective batch size of $16$. We train DreamStyle for $6,000$ and $3,000$ iterations in the CT and SFT stages, respectively, using a LoRA with a rank of $64$ and AdamW~\cite{loshchilov2018decoupled} optimizer with a learning rate of $4\times10^{-5}$.

\subsection{Settings}
For text-guided video stylization, we curate $50$ videos paired with style prompts crafted by a designer as our test set. Since no open-source models specialized in text-guided video stylization are currently available, we opt to compare DreamStyle against three commercial models: Luma~\cite{luma}, Pixverse~\cite{pixverse} and Runway~\cite{runway}. We further expand the aforementioned test set to $90$ videos and $15$ style images (each style image is randomly paired with $6$ videos) to evaluate the style-image-guided task. As a baseline, we select StyleMaster~\cite{ye2025stylemaster}, the only open-source DiT-based method that supports style-image-guided video stylization in an end-to-end manner. For first-frame-guided task, we reuse these $90$ videos and generate stylized first frames for each video using image stylization methods, and then choose VACE~\cite{vace} and VideoX-Fun~\cite{videoxfun} as our competitors.

To evaluate the style consistency, we utilize the CSD~\cite{somepalli2024measuring} score as the quantitative metric. Specifically, the CSD score is computed between the style reference image and each frame of the generated video for style-image-guided task, while for first-frame-guided task, we evaluate this metric between the stylized first frame and all subsequent frames. For text-guided stylization, we employ ViCLIP~\cite{wanginternvid} to measure the similarity between user prompt and stylized video. Moreover, structure preservation is evaluated using the cosine similarity of the patch features (excluding the CLS token) extracted from DINOv2~\cite{oquab2024dinov}. We further assess the overall quality of stylized video with five metrics from VBench~\cite{huang2024vbench}: dynamic degree, image quality, aesthetic quality, subject consistency, and background consistency.

\begin{table}[t]
    \centering
    \small
    \caption{Quantitative comparison. The best and second best results are shown in \textbf{bold} and \underline{underline}.}
    \setlength{\tabcolsep}{4pt}
    \begin{tabular}{c|r|ccccccc}
        \hline
        \multirow{3}{*}{\textbf{Condition}} & \multicolumn{1}{c|}{\multirow{3}{*}{\textbf{Method}}} & \multicolumn{7}{c}{\textbf{Metrics}} \\
        & & CLIP-T / & DINO & Dynamic & Image & Aesthetic & Subject & Background \\
        & & CSD Score & Score & Degree & Quality & Quality & Consistency & Consistency \\
        \hline
        \multirow{4}{*}{Text} & Luma & 0.132 & 0.406 & 0.766 & \underline{0.739} & 0.572 & 0.934 & 0.942 \\
        & Pixverse & \underline{0.155} & 0.451 & 0.766 & \textbf{0.746} & \underline{0.628} & \underline{0.948} & \underline{0.951} \\
        & Runway & 0.154 & \underline{0.504} & \underline{0.809} & 0.725 & 0.606 & 0.940 & 0.944 \\
        & DreamStyle & \textbf{0.167} & \textbf{0.584} & \textbf{0.894} & 0.738 & \textbf{0.656} & \textbf{0.952} & \textbf{0.956} \\
        \hline
        \multirow{3}{*}{Style Image} & StyleMaster (T2V) & 0.198 & - & 0.289 & \textbf{0.723} & 0.610 & 0.936 & 0.935 \\
        & DreamStyle (T2V) & \textbf{0.532} & - & \underline{0.689} & \underline{0.722} & \textbf{0.641} & \textbf{0.950} & \textbf{0.961} \\
        & DreamStyle (V2V) & \underline{0.515} & 0.526 & \textbf{0.867} & 0.704 & \underline{0.635} & \underline{0.938} & \underline{0.948} \\
        \hline
        \multirow{3}{*}{First Frame} & VACE & 0.689 & \textbf{0.716} & \textbf{0.889} & 0.716 & 0.573 & \textbf{0.922} & \underline{0.930} \\
        & VideoX-Fun & \underline{0.766} & \underline{0.702} & 0.844 & \underline{0.726} & \underline{0.594} & 0.915 & 0.924 \\
        & DreamStyle & \textbf{0.851} & 0.640 & \underline{0.856} & \textbf{0.731} & \textbf{0.630} & \underline{0.919} & \textbf{0.932} \\
        \hline
    \end{tabular}
    \label{tab:quant_cmp}
\end{table}

\subsection{Comparisons}
\begin{figure}
    \centering
    \includegraphics[width=0.98\linewidth]{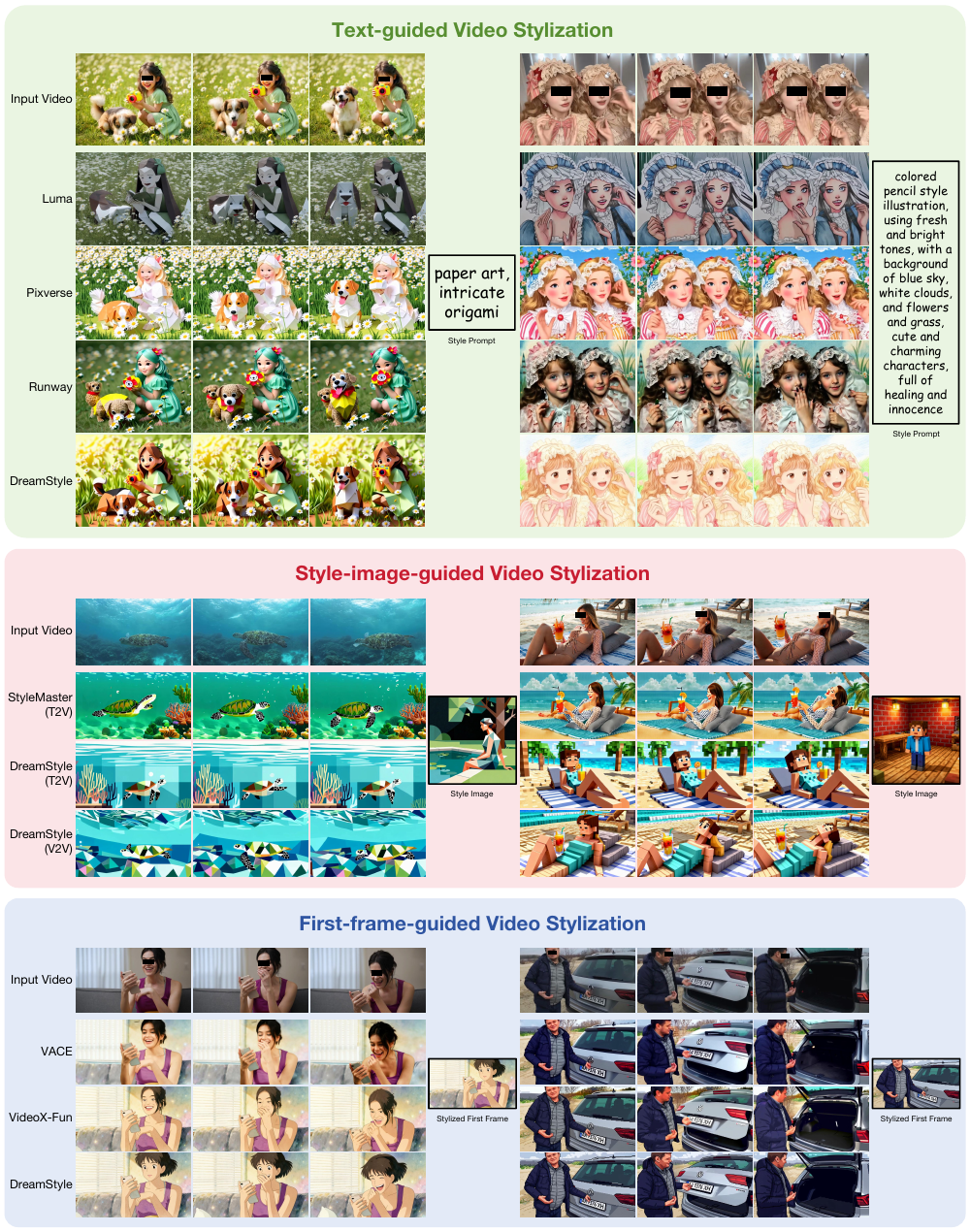}
    \caption{Qualitative comparison on three video stylization tasks.}
    \label{fig:qualitative_cmp}
\end{figure}

\textbf{Quantitative Comparison.}
As shown in Table~\ref{tab:quant_cmp}, we conduct a comprehensive comparison across three video stylization tasks. In text-guided video stylization, DreamStyle achieves the highest CLIP-T (we measure text-video similarity using only the style prompts, thus the CLIP-T is overall lower) and DINO score, indicating that it outperforms the other methods in both style prompt following and structure preservation. This superiority is further evidenced in the visual results in Fig.~\ref{fig:qualitative_cmp}. For the overall video quality assessment, our method also has advantages in most metrics except image quality. Notably, image quality exhibits a negative correlation with dynamic degree, since a high dynamic video tends to involve motion blur, thereby decreasing this metric.
Due to the incomplete open-source of StyleMaster, it supports only T2V instead of V2V stylization. Thus, we include an additional result for DreamStyle in T2V mode, where the video condition is set to empty. Quantitative metrics demonstrate the superior performance of our method, particularly in the aspects of style consistency and dynamic degree. Despite not being explicitly trained for T2V, DreamStyle naturally inherits this capability from the base model thanks to the LoRA training and outperforms its V2V counterpart in most metrics due to fewer constraints.
In first-frame-guided video stylization, DreamStyle presents the optimal style consistency (CSD score) and either the best or second-best video quality metrics. Since the stylized first frames (especially those with geometric deformation) occasionally conflict with the structure of the input video, our method, despite its superior style consistency, is inferior to VACE and VideoX-Fun in structure preservation (DINO score). However, the visual results in Fig.~\ref{fig:qualitative_cmp} confirm that it can still maintain the primary structural elements of the input video.

\noindent\textbf{Qualitative Comparison.}
Fig.~\ref{fig:qualitative_cmp} presents the visual comparisons between our DreamStyle and the competitors. In text-guided video stylization, Luma tends to generate videos with dark tones, and the subject pose, color, and content of its results deviate far from the input videos. Pixverse achieves a higher pose consistency but still suffers from content distortion (e.g., the camera in the left case and the bowknot in the right case). Runway often produces videos with a realistic style bias, failing to accurately render the correct style. By contrast, our DreamStyle not only follows the style prompt but also achieves superior consistency with the input videos in terms of subject pose, color, and content. In style-image-guided video stylization, StyleMaster exhibits limited capability in simple color and texture transfer, while our method can further handle the styles involving geometric shapes. In first-frame-guided video stylization, both VACE and VideoX-Fun struggle to preserve the stylized first frame in the left case. For the right case, although they are able to maintain the major style of the given first frame, serious style degradation occurs in the subsequent frames. By comparison, DreamStyle demonstrates higher style consistency across the stylized first frame, the generated first frame, and all subsequent frames.

\begin{table}[htb]
    \centering
    \small
    \caption{Details of evaluation criteria.}
    \setlength{\tabcolsep}{5pt}
    \resizebox{1.0\textwidth}{!}{
    \begin{tabular}{c|c|l}
        \hline
        \textbf{Metric} & \textbf{Score} & \multicolumn{1}{c}{\textbf{Description}} \\
        \hline
        \multirow{5}{*}{Style Consistency} & 5 & Both the main subject and background perfectly align with the style reference, with stable style throughout the entire video \\
         & 4 & The main subject and background are relatively consistent with the style reference, or there are minor style degradation across the video \\
         & 3 & Either the main subject or the background is somewhat inconsistent with the style reference, or the video exhibits noticeable style variations \\
         & 2 & Neither the main subject nor the background aligns with the style reference, or the video has significant style inconsistencies \\
         & 1 & The main subject and background are completely inconsistent with the style reference \\
        \hline
        \multirow{5}{*}{Content Consistency} & 5 & Both the main subject and background are highly consistent with the input video, and the motion of the main subject is also highly coherent \\
         & 4 & Either the main subject or the background has slight discrepancies from the input video, or the motion of the main subject is somewhat inconsistent \\
         & 3 & Either the main subject or the background has noticeable differences from the input video, or the motion of the main subject is highly inconsistent \\
         & 2 & Both the main subject and background show obvious deviations from the input video \\
         & 1 & The main subject and background are completely unrelated to the input video \\
        \hline
        \multirow{5}{*}{Overall Quality} & 5 & Excellent performance in both style consistency and content consistency, with aesthetically pleasing visuals and rational motion \\
         & 4 & Either style consistency or content consistency needs improvement; or the visuals are generally aesthetically acceptable, with slight motion glitches \\
         & 3 & Either style consistency or content consistency is poor; or the visuals have low aesthetic appeal, with noticeable motion issues \\
         & 2 & Both style consistency and content consistency are poor, with unappealing visuals and significant motion issues \\
         & 1 & Extremely poor performance in both style consistency and content consistency \\
        \hline
    \end{tabular}}
    \label{tab:eval_criteria}
\end{table}

\subsection{User Study}
Human feedback serves as an important method for evaluating stylization performance, thus we conduct a user study focusing on three core metrics: style consistency, content consistency, and overall quality. Each metric is rated on a 1-5 scale, with the detailed evaluation criteria provided in Table~\ref{tab:eval_criteria}. We recruited 20 professional data annotators as evaluators and randomly selected 10, 20, and 20 samples from the text-guided, style-image-guided, and first-frame-guided test sets, respectively, for blind evaluation. As shown in Table~\ref{tab:user_study}, DreamStyle outperforms other methods across all three stylization tasks, with a notable superiority in style consistency. Its overall quality score reaches approximately 4 or higher, reflecting user recognition of its performance.

\begin{table}[h]
    \centering
    \caption{User study on three video stylization tasks.}
    \begin{tabular}{c|r|ccc}
        \hline
        \multirow{3}{*}{\textbf{Condition}} & \multicolumn{1}{c|}{\multirow{3}{*}{\textbf{Method}}} & \multicolumn{3}{c}{\textbf{Metrics}} \\
         & & Style & Content & Overall \\
         & & Consistency & Consistency & Quality \\
        \hline
        \multirow{4}{*}{Text} & Luma & 2.05 & 2.58 & 2.24 \\
        & Pixverse & 2.83 & 2.95 & 2.82 \\
        & Runway & 2.52 & 2.80 & 2.59 \\
        & DreamStyle & \textbf{4.14} & \textbf{3.95} & \textbf{3.95} \\
        \hline
        \multirow{2}{*}{Style Image} & StyleMaster & 1.17 & - & 1.31 \\
        & DreamStyle & \textbf{4.36} & 3.87 & \textbf{4.20} \\
        \hline
        \multirow{3}{*}{First Frame} & VACE & 2.35 & \textbf{4.30} & 2.79 \\
        & VideoX-Fun & 3.19 & 4.22 & 3.42 \\
        & DreamStyle & \textbf{4.37} & 4.12 & \textbf{4.24} \\
        \hline
    \end{tabular}
    \label{tab:user_study}
\end{table}

\subsection{Extended Applications}
\begin{figure}
    \centering
    \includegraphics[width=0.93\linewidth]{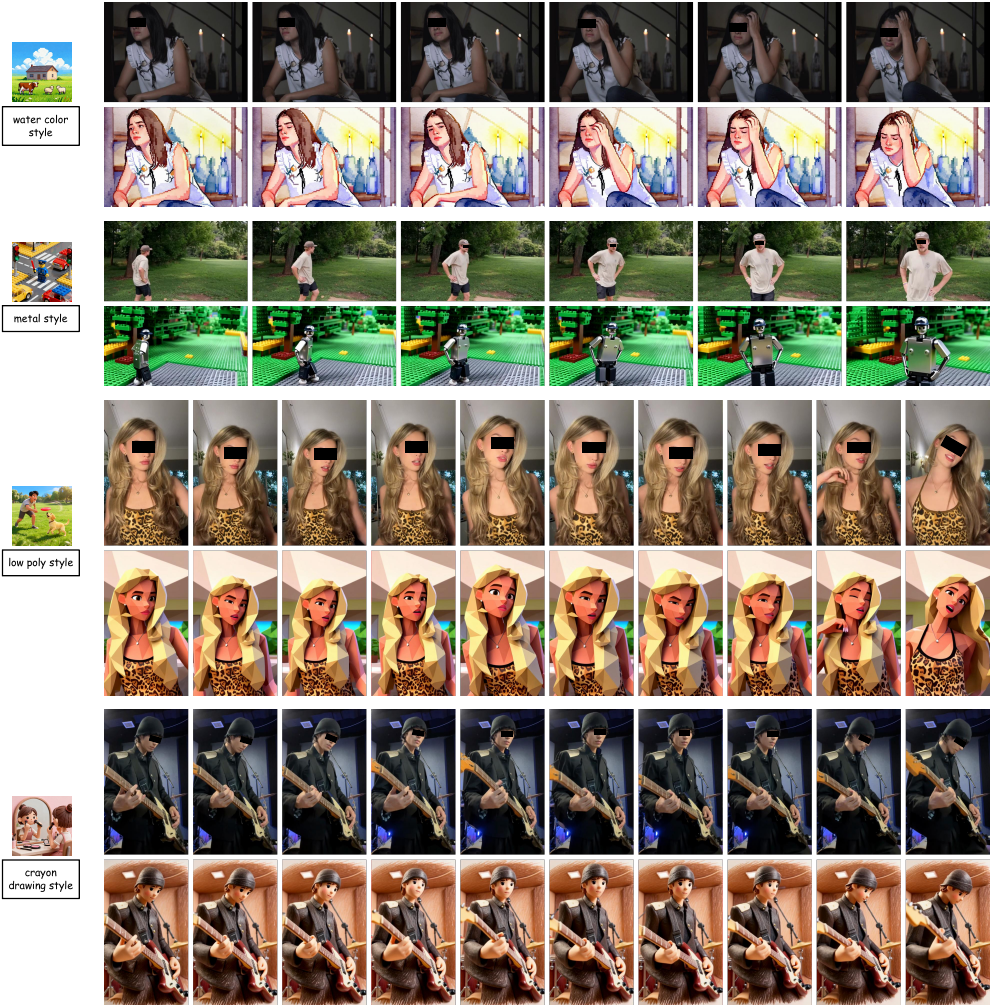}
    \caption{Visual results of multi-style fusion.}
    \label{fig:style_fusion}
\end{figure}

Although DreamStyle is trained with only a single condition type at a time, it still supports multiple guidance modalities during inference, thereby unlocking its potential to enable broader extended applications. Below, we highlight two representative scenarios:

\noindent\textbf{Multi-Style Fusion.}
As shown in Fig.~\ref{fig:style_fusion}, DreamStyle can naturally integrates the style cues from both text prompts and style images, demonstrating its capability to fuse diverse style references and create a novel style. This flexibility allows for a creative combination of abstract textual description and precise visual reference, exhibiting the potential beyond single guidance.

\begin{figure}
    \centering
    \includegraphics[width=1.0\linewidth]{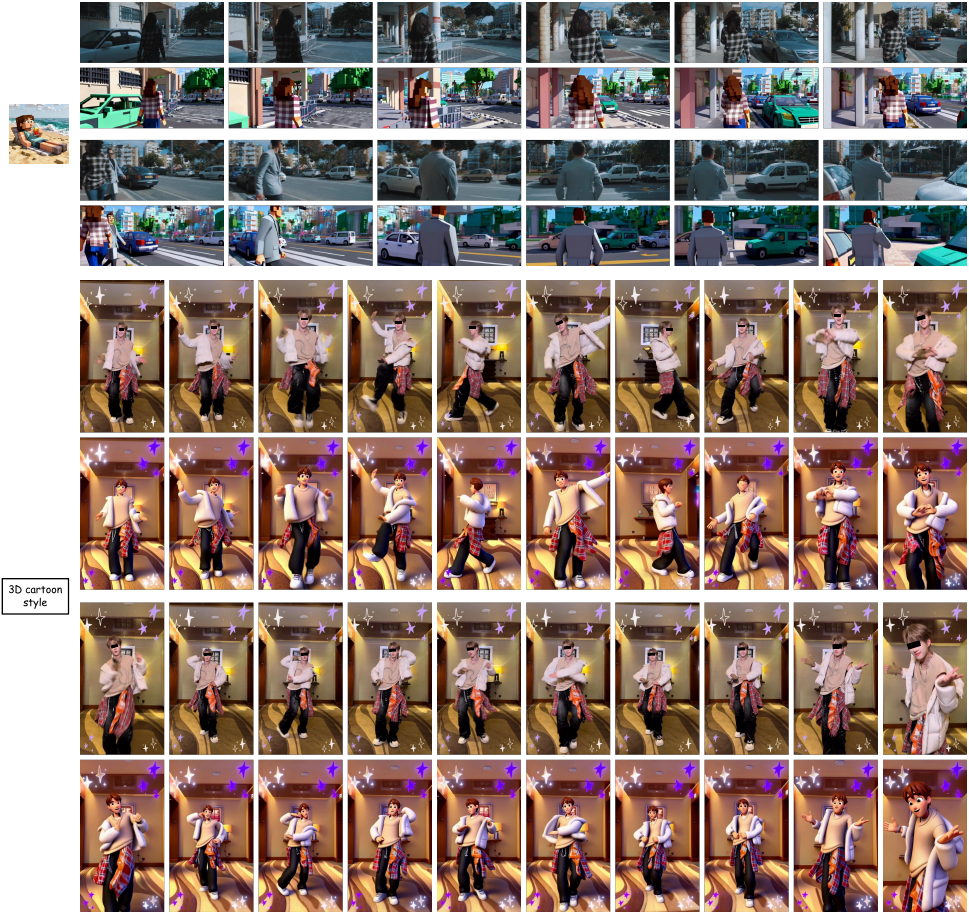}
    \caption{Visual results of long-video stylization.}
    \label{fig:long_video}
\end{figure}

\noindent\textbf{Long-Video Stylization.}
By leveraging the last frame of a generated short video as the first frame condition for the next segment, we can seamlessly concatenate two short video clips. Thus, a combination of first frame guidance and text or style image enables DreamStyle to overcome the 5-second duration limit, supporting stylization for longer video sequences (except multi-shot video due to the inherent limitations of the base model and training data). Fig.~\ref{fig:long_video} presents two long-video stylization examples, guided by style image and text, respectively.

\begin{table}[htb]
    \centering
    \caption{Quantitative comparison of ablation studies.}
    \begin{tabular}{r|cc}
        \hline
         & CSD Score & DINO Score \\
         \hline
         w.o. Token-specific LoRA & 0.413 & 0.518 \\
         Only CT Data & 0.459 & \textbf{0.547} \\
         Only SFT Data & \textbf{0.535} & 0.483 \\
         Full & \underline{0.515} & \underline{0.526} \\
         \hline
    \end{tabular}
    \label{tab:ab_cmp}
\end{table}

\begin{figure}
    \centering
    \includegraphics[width=0.75\linewidth]{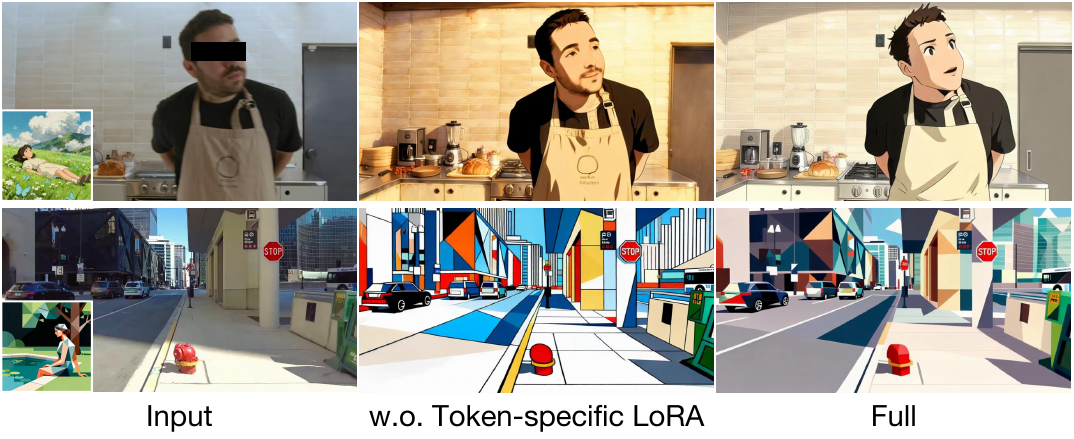}
    \caption{Impact of token-specific LoRA.}
    \label{fig:ab_lora}
\end{figure}

\begin{figure}
    \centering
    \includegraphics[width=0.8\linewidth]{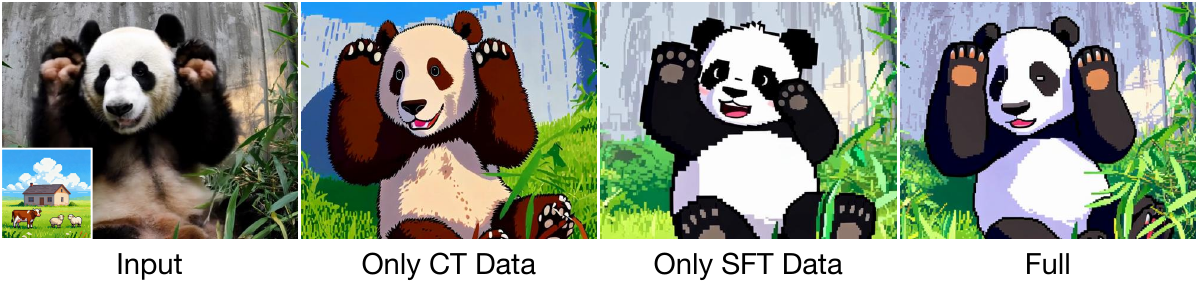}
    \caption{Visual comparison across different datasets.}
    \label{fig:ab_data}
\end{figure}

\subsection{Ablation Studies}
\textbf{Token-specific LoRA.}
The proposed token-specific LoRA plays a critical role in mitigating interference among distinct condition tokens. To validate this, we design an ablation experiment where DreamStyle is trained with a standard LoRA—here, different condition tokens are distinguished solely through frame positions and mask values ($1$ for first-frame tokens, $0$ for video tokens, and $-1$ for style-image tokens). We focus on the style-image-guided stylization task for evaluation. As shown in Table~\ref{tab:ab_cmp}, the standard LoRA exhibits a significantly negative influence on style consistency (CSD score) and slightly reduces structure preservation (DINO score). Visual evidence in Fig.~\ref{fig:ab_lora} further indicates this point, where the problems of style degradation (first row) and style confusion (second row) arise in the absence of token-specific LoRA.

\noindent\textbf{Datasets.}
To validate the necessity of both datasets (with distinct scales and quality) and two-stage training, we conduct an ablation experiment where DreamStyle is trained on only the CT dataset, only the SFT dataset, and both of them in two stages. Due to the limited quality and style consistency of the CT dataset, training exclusively on it yields suboptimal stylization performance. Both the quantitative metric (lower CSD score) in Table~\ref{tab:ab_cmp} and the visual result (failure to render the pixel pattern) in Fig.~\ref{fig:ab_data} confirm this shortcoming. Conversely, the SFT dataset contains manually filtered paired videos with high quality and strong style consistency, but its limited size makes it insufficient to adapt the I2V base model into a robust V2V model for stylization (particularly, there is no strict alignment between the paired videos due to the existence of geometric deformation). Thus, as shown in Table~\ref{tab:ab_cmp}, training solely on it, despite achieving the best CSD score, exhibits the worst performance on structure preservation, which is also evidenced in Fig.~\ref{fig:ab_data} (the pose of the stylized panda differs from the input). As expected, training on the CT and SFT datasets in turn achieves a robust balance between style consistency and structure preservation.

%% file: sections/5_conclusion.tex
\section{Conclusion}
\label{sec:conclusion}
In this paper, we propose DreamStyle, the first unified framework for video stylization that supports three style conditions: text, style image, and first frame. Recognizing the critical role of high-quality paired video datasets in training, we develop a systematic data curation pipeline consisting of two key steps: (1) leveraging the SOTA image stylization models to obtain the stylized first frame; (2) animating the raw and stylized first frames using an I2V model equipped with ControlNets. In each step, we further apply automatic and manual filtering to ensure data quality. DreamStyle is built upon an I2V model, which can be efficiently extended to the V2V model without involving too much extra computation overhead. Moreover, to address the inter-token confusion among different style conditions within a unified model, we introduce a novel token-specific LoRA module. With our high-quality dataset, well-designed model architecture, and two-stage training strategy, DreamStyle archives competitive performance on various video stylization tasks.

%% file: main.bib
@String(TOG= {ACM Trans. Graph.})

@String(ICLR = {Int. Conf. Learn. Represent.})

@String(TOG   = {ACM TOG})

@String(ICLR  = {ICLR})

@inproceedings{ye2025stylemaster,
  title={Stylemaster: Stylize your video with artistic generation and translation},
  author={Ye, Zixuan and Huang, Huijuan and Wang, Xintao and Wan, Pengfei and Zhang, Di and Luo, Wenhan},
  booktitle={Proceedings of the Computer Vision and Pattern Recognition Conference},
  pages={2630--2640},
  year={2025}
}

@inproceedings{wang2025omnistyle,
  title={OmniStyle: Filtering High Quality Style Transfer Data at Scale},
  author={Wang, Ye and Liu, Ruiqi and Lin, Jiang and Liu, Fei and Yi, Zili and Wang, Yilin and Ma, Rui},
  booktitle={Proceedings of the Computer Vision and Pattern Recognition Conference},
  pages={7847--7856},
  year={2025}
}

@article{liu2024stylecrafter,
  title={Stylecrafter: Taming artistic video diffusion with reference-augmented adapter learning},
  author={Liu, Gongye and Xia, Menghan and Zhang, Yong and Chen, Haoxin and Xing, Jinbo and Wang, Yibo and Wang, Xintao and Shan, Ying and Yang, Yujiu},
  journal={ACM Transactions on Graphics (TOG)},
  volume={43},
  number={6},
  pages={1--10},
  year={2024},
  publisher={ACM New York, NY, USA}
}

@article{chefer2024still,
  title={Still-moving: Customized video generation without customized video data},
  author={Chefer, Hila and Zada, Shiran and Paiss, Roni and Ephrat, Ariel and Tov, Omer and Rubinstein, Michael and Wolf, Lior and Dekel, Tali and Michaeli, Tomer and Mosseri, Inbar},
  journal={ACM Transactions on Graphics (TOG)},
  volume={43},
  number={6},
  pages={1--11},
  year={2024},
  publisher={ACM New York, NY, USA}
}

@article{song2024univst,
  title={Univst: A unified framework for training-free localized video style transfer},
  author={Song, Quanjian and Lin, Mingbao and Zhan, Wengyi and Yan, Shuicheng and Cao, Liujuan and Ji, Rongrong},
  journal={arXiv preprint arXiv:2410.20084},
  year={2024}
}

@inproceedings{
geyer2024tokenflow,
title={TokenFlow: Consistent Diffusion Features for Consistent Video Editing},
author={Michal Geyer and Omer Bar-Tal and Shai Bagon and Tali Dekel},
booktitle={The Twelfth International Conference on Learning Representations},
year={2024},
url={https://openreview.net/forum?id=lKK50q2MtV}
}

@article{
ku2024anyvv,
title={AnyV2V: A Tuning-Free Framework For Any Video-to-Video Editing Tasks},
author={Max Ku and Cong Wei and Weiming Ren and Huan Yang and Wenhu Chen},
journal={Transactions on Machine Learning Research},
issn={2835-8856},
year={2024},
url={https://openreview.net/forum?id=RFrJCkw2oa},
note={Reproducibility Certification}
}

@inproceedings{li2024styletokenizer,
  title={Styletokenizer: Defining image style by a single instance for controlling diffusion models},
  author={Li, Wen and Fang, Muyuan and Zou, Cheng and Gong, Biao and Zheng, Ruobing and Wang, Meng and Chen, Jingdong and Yang, Ming},
  booktitle={European Conference on Computer Vision},
  pages={110--126},
  year={2024},
  organization={Springer}
}

@inproceedings{peebles2023scalable,
  title={Scalable diffusion models with transformers},
  author={Peebles, William and Xie, Saining},
  booktitle={Proceedings of the IEEE/CVF international conference on computer vision},
  pages={4195--4205},
  year={2023}
}

@inproceedings{radford2021learning,
  title={Learning transferable visual models from natural language supervision},
  author={Radford, Alec and Kim, Jong Wook and Hallacy, Chris and Ramesh, Aditya and Goh, Gabriel and Agarwal, Sandhini and Sastry, Girish and Askell, Amanda and Mishkin, Pamela and Clark, Jack and others},
  booktitle={International conference on machine learning},
  pages={8748--8763},
  year={2021},
  organization={PmLR}
}

@inproceedings{gatys2016image,
  title={Image style transfer using convolutional neural networks},
  author={Gatys, Leon A and Ecker, Alexander S and Bethge, Matthias},
  booktitle={Proceedings of the IEEE conference on computer vision and pattern recognition},
  pages={2414--2423},
  year={2016}
}

@inproceedings{simonyan2015very,
  title={Very deep convolutional networks for large-scale image recognition},
  author={Simonyan, K and Zisserman, A},
  booktitle={3rd International Conference on Learning Representations (ICLR 2015)},
  year={2015},
  organization={Computational and Biological Learning Society}
}

@inproceedings{
song2021denoising,
title={Denoising Diffusion Implicit Models},
author={Jiaming Song and Chenlin Meng and Stefano Ermon},
booktitle={International Conference on Learning Representations},
year={2021},
url={https://openreview.net/forum?id=St1giarCHLP}
}

@article{ho2020denoising,
  title={Denoising diffusion probabilistic models},
  author={Ho, Jonathan and Jain, Ajay and Abbeel, Pieter},
  journal={Advances in neural information processing systems},
  volume={33},
  pages={6840--6851},
  year={2020}
}

@inproceedings{nichol2021improved,
  title={Improved denoising diffusion probabilistic models},
  author={Nichol, Alexander Quinn and Dhariwal, Prafulla},
  booktitle={International conference on machine learning},
  pages={8162--8171},
  year={2021},
  organization={PMLR}
}

@article{dhariwal2021diffusion,
  title={Diffusion models beat gans on image synthesis},
  author={Dhariwal, Prafulla and Nichol, Alexander},
  journal={Advances in neural information processing systems},
  volume={34},
  pages={8780--8794},
  year={2021}
}

@article{saharia2022photorealistic,
  title={Photorealistic text-to-image diffusion models with deep language understanding},
  author={Saharia, Chitwan and Chan, William and Saxena, Saurabh and Li, Lala and Whang, Jay and Denton, Emily L and Ghasemipour, Kamyar and Gontijo Lopes, Raphael and Karagol Ayan, Burcu and Salimans, Tim and others},
  journal={Advances in neural information processing systems},
  volume={35},
  pages={36479--36494},
  year={2022}
}

@inproceedings{rombach2022high,
  title={High-resolution image synthesis with latent diffusion models},
  author={Rombach, Robin and Blattmann, Andreas and Lorenz, Dominik and Esser, Patrick and Ommer, Bj{\"o}rn},
  booktitle={Proceedings of the IEEE/CVF conference on computer vision and pattern recognition},
  pages={10684--10695},
  year={2022}
}

@inproceedings{
podell2024sdxl,
title={{SDXL}: Improving Latent Diffusion Models for High-Resolution Image Synthesis},
author={Dustin Podell and Zion English and Kyle Lacey and Andreas Blattmann and Tim Dockhorn and Jonas M{\"u}ller and Joe Penna and Robin Rombach},
booktitle={The Twelfth International Conference on Learning Representations},
year={2024},
url={https://openreview.net/forum?id=di52zR8xgf}
}

@inproceedings{Kingma2014,
  author = {Kingma, Diederik P. and Welling, Max},
  booktitle = {2nd International Conference on Learning Representations, {ICLR} 2014, Banff, AB, Canada, April 14-16, 2014, Conference Track Proceedings},
  title = {{Auto-Encoding Variational Bayes}},
  year = 2014
}

@article{blattmann2023stable,
  title={Stable video diffusion: Scaling latent video diffusion models to large datasets},
  author={Blattmann, Andreas and Dockhorn, Tim and Kulal, Sumith and Mendelevitch, Daniel and Kilian, Maciej and Lorenz, Dominik and Levi, Yam and English, Zion and Voleti, Vikram and Letts, Adam and others},
  journal={arXiv preprint arXiv:2311.15127},
  year={2023}
}

@inproceedings{blattmann2023align,
  title={Align your latents: High-resolution video synthesis with latent diffusion models},
  author={Blattmann, Andreas and Rombach, Robin and Ling, Huan and Dockhorn, Tim and Kim, Seung Wook and Fidler, Sanja and Kreis, Karsten},
  booktitle={Proceedings of the IEEE/CVF conference on computer vision and pattern recognition},
  pages={22563--22575},
  year={2023}
}

@inproceedings{
guo2024animatediff,
title={AnimateDiff: Animate Your Personalized Text-to-Image Diffusion Models without Specific Tuning},
author={Yuwei Guo and Ceyuan Yang and Anyi Rao and Zhengyang Liang and Yaohui Wang and Yu Qiao and Maneesh Agrawala and Dahua Lin and Bo Dai},
booktitle={The Twelfth International Conference on Learning Representations},
year={2024},
url={https://openreview.net/forum?id=Fx2SbBgcte}
}

@inproceedings{guo2024i2v,
  title={I2v-adapter: A general image-to-video adapter for diffusion models},
  author={Guo, Xun and Zheng, Mingwu and Hou, Liang and Gao, Yuan and Deng, Yufan and Wan, Pengfei and Zhang, Di and Liu, Yufan and Hu, Weiming and Zha, Zhengjun and others},
  booktitle={ACM SIGGRAPH 2024 Conference Papers},
  pages={1--12},
  year={2024}
}

@inproceedings{ronneberger2015u,
  title={U-net: Convolutional networks for biomedical image segmentation},
  author={Ronneberger, Olaf and Fischer, Philipp and Brox, Thomas},
  booktitle={International Conference on Medical image computing and computer-assisted intervention},
  pages={234--241},
  year={2015},
  organization={Springer}
}

@article{brooks2024video,
  title={Video generation models as world simulators},
  author={Brooks, Tim and Peebles, Bill and Holmes, Connor and DePue, Will and Guo, Yufei and Jing, Li and Schnurr, David and Taylor, Joe and Luhman, Troy and Luhman, Eric and others},
  journal={OpenAI Blog},
  volume={1},
  number={8},
  pages={1},
  year={2024}
}

@inproceedings{
hong2023cogvideo,
title={CogVideo: Large-scale Pretraining for Text-to-Video Generation via Transformers},
author={Wenyi Hong and Ming Ding and Wendi Zheng and Xinghan Liu and Jie Tang},
booktitle={The Eleventh International Conference on Learning Representations },
year={2023},
url={https://openreview.net/forum?id=rB6TpjAuSRy}
}

@inproceedings{
yang2025cogvideox,
title={CogVideoX: Text-to-Video Diffusion Models with An Expert Transformer},
author={Zhuoyi Yang and Jiayan Teng and Wendi Zheng and Ming Ding and Shiyu Huang and Jiazheng Xu and Yuanming Yang and Wenyi Hong and Xiaohan Zhang and Guanyu Feng and Da Yin and Yuxuan.Zhang and Weihan Wang and Yean Cheng and Bin Xu and Xiaotao Gu and Yuxiao Dong and Jie Tang},
booktitle={The Thirteenth International Conference on Learning Representations},
year={2025},
url={https://openreview.net/forum?id=LQzN6TRFg9}
}

@article{wan2025wan,
  title={Wan: Open and advanced large-scale video generative models},
  author={Wan, Team and Wang, Ang and Ai, Baole and Wen, Bin and Mao, Chaojie and Xie, Chen-Wei and Chen, Di and Yu, Feiwu and Zhao, Haiming and Yang, Jianxiao and others},
  journal={arXiv preprint arXiv:2503.20314},
  year={2025}
}

@article{gao2025seedance,
  title={Seedance 1.0: Exploring the Boundaries of Video Generation Models},
  author={Gao, Yu and Guo, Haoyuan and Hoang, Tuyen and Huang, Weilin and Jiang, Lu and Kong, Fangyuan and Li, Huixia and Li, Jiashi and Li, Liang and Li, Xiaojie and others},
  journal={arXiv preprint arXiv:2506.09113},
  year={2025}
}

@inproceedings{chen2017stylebank,
  title={Stylebank: An explicit representation for neural image style transfer},
  author={Chen, Dongdong and Yuan, Lu and Liao, Jing and Yu, Nenghai and Hua, Gang},
  booktitle={Proceedings of the IEEE conference on computer vision and pattern recognition},
  pages={1897--1906},
  year={2017}
}

@inproceedings{huang2017arbitrary,
  title={Arbitrary style transfer in real-time with adaptive instance normalization},
  author={Huang, Xun and Belongie, Serge},
  booktitle={Proceedings of the IEEE international conference on computer vision},
  pages={1501--1510},
  year={2017}
}

@article{risser2017stable,
  title={Stable and controllable neural texture synthesis and style transfer using histogram losses},
  author={Risser, Eric and Wilmot, Pierre and Barnes, Connelly},
  journal={arXiv preprint arXiv:1701.08893},
  year={2017}
}

@article{ye2023ip,
  title={Ip-adapter: Text compatible image prompt adapter for text-to-image diffusion models},
  author={Ye, Hu and Zhang, Jun and Liu, Sibo and Han, Xiao and Yang, Wei},
  journal={arXiv preprint arXiv:2308.06721},
  year={2023}
}

@article{wang2024instantstyle,
  title={Instantstyle: Free lunch towards style-preserving in text-to-image generation},
  author={Wang, Haofan and Spinelli, Matteo and Wang, Qixun and Bai, Xu and Qin, Zekui and Chen, Anthony},
  journal={arXiv preprint arXiv:2404.02733},
  year={2024}
}

@article{xing2024csgo,
  title={Csgo: Content-style composition in text-to-image generation},
  author={Xing, Peng and Wang, Haofan and Sun, Yanpeng and Wang, Qixun and Bai, Xu and Ai, Hao and Huang, Renyuan and Li, Zechao},
  journal={arXiv preprint arXiv:2408.16766},
  year={2024}
}

@inproceedings{qi2024deadiff,
  title={Deadiff: An efficient stylization diffusion model with disentangled representations},
  author={Qi, Tianhao and Fang, Shancheng and Wu, Yanze and Xie, Hongtao and Liu, Jiawei and Chen, Lang and He, Qian and Zhang, Yongdong},
  booktitle={Proceedings of the IEEE/CVF conference on computer vision and pattern recognition},
  pages={8693--8702},
  year={2024}
}

@inproceedings{
hu2022lora,
title={Lo{RA}: Low-Rank Adaptation of Large Language Models},
author={Edward J Hu and yelong shen and Phillip Wallis and Zeyuan Allen-Zhu and Yuanzhi Li and Shean Wang and Lu Wang and Weizhu Chen},
booktitle={International Conference on Learning Representations},
year={2022},
url={https://openreview.net/forum?id=nZeVKeeFYf9}
}

@inproceedings{vace,
    title = {VACE: All-in-One Video Creation and Editing},
    author = {Jiang, Zeyinzi and Han, Zhen and Mao, Chaojie and Zhang, Jingfeng and Pan, Yulin and Liu, Yu},
    booktitle = {Proceedings of the IEEE/CVF International Conference on Computer Vision},
    pages = {17191-17202},
    year = {2025}
}

@article{seedream2025seedream,
  title={Seedream 4.0: Toward next-generation multimodal image generation},
  author={Seedream, Team and Chen, Yunpeng and Gao, Yu and Gong, Lixue and Guo, Meng and Guo, Qiushan and Guo, Zhiyao and Hou, Xiaoxia and Huang, Weilin and Huang, Yixuan and others},
  journal={arXiv preprint arXiv:2509.20427},
  year={2025}
}

@article{tian2024hydralora,
  title={Hydralora: An asymmetric lora architecture for efficient fine-tuning},
  author={Tian, Chunlin and Shi, Zhan and Guo, Zhijiang and Li, Li and Xu, Cheng-Zhong},
  journal={Advances in Neural Information Processing Systems},
  volume={37},
  pages={9565--9584},
  year={2024}
}

@software{videoxfun,
title = {A more flexible framework that can generate videos at any resolution and creates videos from images},
url = {https://github.com/aigc-apps/VideoX-Fun},
years = {2025}
}

@software{luma,
title = {Luma AI},
url = {https://lumalabs.ai},
years = {2025}
}

@software{pixverse,
title = {PixVerse AI Video Generator},
url = {https://app.pixverse.ai},
years = {2025}
}

@software{runway,
title = {Runway AI Image and Video Generator},
url = {https://runwayml.com},
years = {2025}
}

@article{
oquab2024dinov,
title={{DINO}v2: Learning Robust Visual Features without Supervision},
author={Maxime Oquab and Timoth{\'e}e Darcet and Th{\'e}o Moutakanni and Huy V. Vo and Marc Szafraniec and Vasil Khalidov and Pierre Fernandez and Daniel HAZIZA and Francisco Massa and Alaaeldin El-Nouby and Mido Assran and Nicolas Ballas and Wojciech Galuba and Russell Howes and Po-Yao Huang and Shang-Wen Li and Ishan Misra and Michael Rabbat and Vasu Sharma and Gabriel Synnaeve and Hu Xu and Herve Jegou and Julien Mairal and Patrick Labatut and Armand Joulin and Piotr Bojanowski},
journal={Transactions on Machine Learning Research},
issn={2835-8856},
year={2024},
url={https://openreview.net/forum?id=a68SUt6zFt},
note={Featured Certification}
}

@inproceedings{huang2024vbench,
  title={Vbench: Comprehensive benchmark suite for video generative models},
  author={Huang, Ziqi and He, Yinan and Yu, Jiashuo and Zhang, Fan and Si, Chenyang and Jiang, Yuming and Zhang, Yuanhan and Wu, Tianxing and Jin, Qingyang and Chanpaisit, Nattapol and others},
  booktitle={Proceedings of the IEEE/CVF Conference on Computer Vision and Pattern Recognition},
  pages={21807--21818},
  year={2024}
}

@article{somepalli2024measuring,
  title={Measuring style similarity in diffusion models},
  author={Somepalli, Gowthami and Gupta, Anubhav and Gupta, Kamal and Palta, Shramay and Goldblum, Micah and Geiping, Jonas and Shrivastava, Abhinav and Goldstein, Tom},
  journal={arXiv preprint arXiv:2404.01292},
  year={2024}
}

@article{ye2025unic,
  title={UNIC: Unified In-Context Video Editing},
  author={Ye, Zixuan and He, Xuanhua and Liu, Quande and Wang, Qiulin and Wang, Xintao and Wan, Pengfei and Zhang, Di and Gai, Kun and Chen, Qifeng and Luo, Wenhan},
  journal={arXiv preprint arXiv:2506.04216},
  year={2025}
}

@article{guo2024pulid,
  title={Pulid: Pure and lightning id customization via contrastive alignment},
  author={Guo, Zinan and Wu, Yanze and Zhuowei, Chen and Zhang, Peng and He, Qian and others},
  journal={Advances in neural information processing systems},
  volume={37},
  pages={36777--36804},
  year={2024}
}

@inproceedings{zhang2023adding,
  title={Adding conditional control to text-to-image diffusion models},
  author={Zhang, Lvmin and Rao, Anyi and Agrawala, Maneesh},
  booktitle={Proceedings of the IEEE/CVF international conference on computer vision},
  pages={3836--3847},
  year={2023}
}

@article{zhang2024vision,
  title={Vision-language models for vision tasks: A survey},
  author={Zhang, Jingyi and Huang, Jiaxing and Jin, Sheng and Lu, Shijian},
  journal={IEEE transactions on pattern analysis and machine intelligence},
  volume={46},
  number={8},
  pages={5625--5644},
  year={2024},
  publisher={IEEE}
}

@inproceedings{
loshchilov2018decoupled,
title={Decoupled Weight Decay Regularization},
author={Ilya Loshchilov and Frank Hutter},
booktitle={International Conference on Learning Representations},
year={2019},
url={https://openreview.net/forum?id=Bkg6RiCqY7},
}

@inproceedings{sohl2015deep,
  title={Deep unsupervised learning using nonequilibrium thermodynamics},
  author={Sohl-Dickstein, Jascha and Weiss, Eric and Maheswaranathan, Niru and Ganguli, Surya},
  booktitle={International conference on machine learning},
  pages={2256--2265},
  year={2015},
  organization={pmlr}
}

@article{kong2024hunyuanvideo,
  title={Hunyuanvideo: A systematic framework for large video generative models},
  author={Kong, Weijie and Tian, Qi and Zhang, Zijian and Min, Rox and Dai, Zuozhuo and Zhou, Jin and Xiong, Jiangfeng and Li, Xin and Wu, Bo and Zhang, Jianwei and others},
  journal={arXiv preprint arXiv:2412.03603},
  year={2024}
}

@inproceedings{wanginternvid,
  title={InternVid: A Large-scale Video-Text Dataset for Multimodal Understanding and Generation},
  author={Wang, Yi and He, Yinan and Li, Yizhuo and Li, Kunchang and Yu, Jiashuo and Ma, Xin and Li, Xinhao and Chen, Guo and Chen, Xinyuan and Wang, Yaohui and others},
  booktitle={The Twelfth International Conference on Learning Representations},
  year={2024}
}

@inproceedings{lipmanflow,
  title={Flow Matching for Generative Modeling},
  author={Lipman, Yaron and Chen, Ricky TQ and Ben-Hamu, Heli and Nickel, Maximilian and Le, Matthew},
  booktitle={The Eleventh International Conference on Learning Representations},
  year={2023}
}

@inproceedings{dou2024loramoe,
  title={LoRAMoE: Alleviating world knowledge forgetting in large language models via MoE-style plugin},
  author={Dou, Shihan and Zhou, Enyu and Liu, Yan and Gao, Songyang and Shen, Wei and Xiong, Limao and Zhou, Yuhao and Wang, Xiao and Xi, Zhiheng and Fan, Xiaoran and others},
  booktitle={Proceedings of the 62nd Annual Meeting of the Association for Computational Linguistics (Volume 1: Long Papers)},
  pages={1932--1945},
  year={2024}
}
